\documentclass[sigconf]{acmart}
\usepackage{multirow}
\usepackage{algorithm}
\usepackage{algorithmic}
\usepackage{multirow}
\usepackage{xcolor}
\usepackage{svg}
\usepackage{bbding}

\newtheorem{Definition}{Definition}
\AtBeginDocument{%
\providecommand\BibTeX{{%
\normalfont B\kern-0.5em{\scshape i\kern-0.25em b}\kern-0.8em\TeX}}}

\copyrightyear{2025}
\acmYear{2025}
\setcopyright{acmlicensed}\acmConference[MM '25]{Proceedings of the 33rd ACM International Conference on Multimedia}{October 27--31, 2025}{Dublin, Ireland}
\acmBooktitle{Proceedings of the 33rd ACM International Conference on Multimedia (MM '25), October 27--31, 2025, Dublin, Ireland}
\acmDOI{10.1145/3746027.3755033}
\acmISBN{979-8-4007-2035-2/2025/10}

\begin{document}

\title{Enhancing  Multi-view  Open-set Learning via Ambiguity Uncertainty Calibration  and View-wise Debiasing}
 \author{Zihan Fang}

 \affiliation{%
   \institution{Fuzhou University}
   \city{Fuzhou}
   \country{China}
 }
 \email{fzihan11@163.com}

 \author{Zhiyong Xu}

 \affiliation{%
   \institution{Fuzhou University}
   \city{Fuzhou}
   \country{China}
 }
 \email{teddy\_xu@163.com}
 \author{Lan Du}

 \affiliation{%
   \institution{Monash University}
   \city{Melbourne}
   \country{Australia}
 }
 \email{lan.du@monash.edu}
 \author{Shide Du}

 \affiliation{%
   \institution{Fuzhou University}
   \city{Fuzhou}
   \country{China}
 }
 \email{dushidems@gmail.com}
 \author{Zhiling Cai}

 \affiliation{%
   \institution{Fujian Agriculture and Forestry University}
   \city{Fuzhou}
   \country{China}
 }
 \email{zhilingcai@126.com}

 \author{Shiping Wang}
\authornote{Corresponding author}

 \affiliation{%
   \institution{Fuzhou University}
   \city{Fuzhou}
   \country{China}
 }
 \email{shipingwangphd@163.com}
\renewcommand{\shortauthors}{Zihan Fang et al.}

\begin{abstract}
Existing multi-view learning models struggle in open-set scenarios due to their implicit assumption of class completeness.
Moreover, static view-induced biases, which arise from spurious view-label associations formed during training, further degrade their ability to recognize unknown categories.
In this paper, we propose a multi-view open-set learning framework via ambiguity uncertainty calibration and view-wise debiasing.
To simulate ambiguous samples, we design \textit{O}-Mix, a novel synthesis strategy to generate virtual samples with calibrated open-set ambiguity uncertainty.
These samples are further processed by an auxiliary ambiguity perception network that captures atypical patterns for improved open-set adaptation.
Furthermore, we incorporate an HSIC-based contrastive debiasing module that enforces independence between view-specific ambiguous and view-consistent representations, encouraging the model to learn generalizable features.
Extensive experiments on diverse multi-view benchmarks demonstrate that the proposed framework consistently enhances unknown-class recognition while preserving strong closed-set performance.
The source code are available.\footnote{\scriptsize\url{https://github.com/ZihanFang11/2025\_MOCD\_ACMMM}}
\end{abstract}



\maketitle

\section{Introduction}
\begin{figure}[t]
  \centering
  \includegraphics[width=0.47\textwidth]{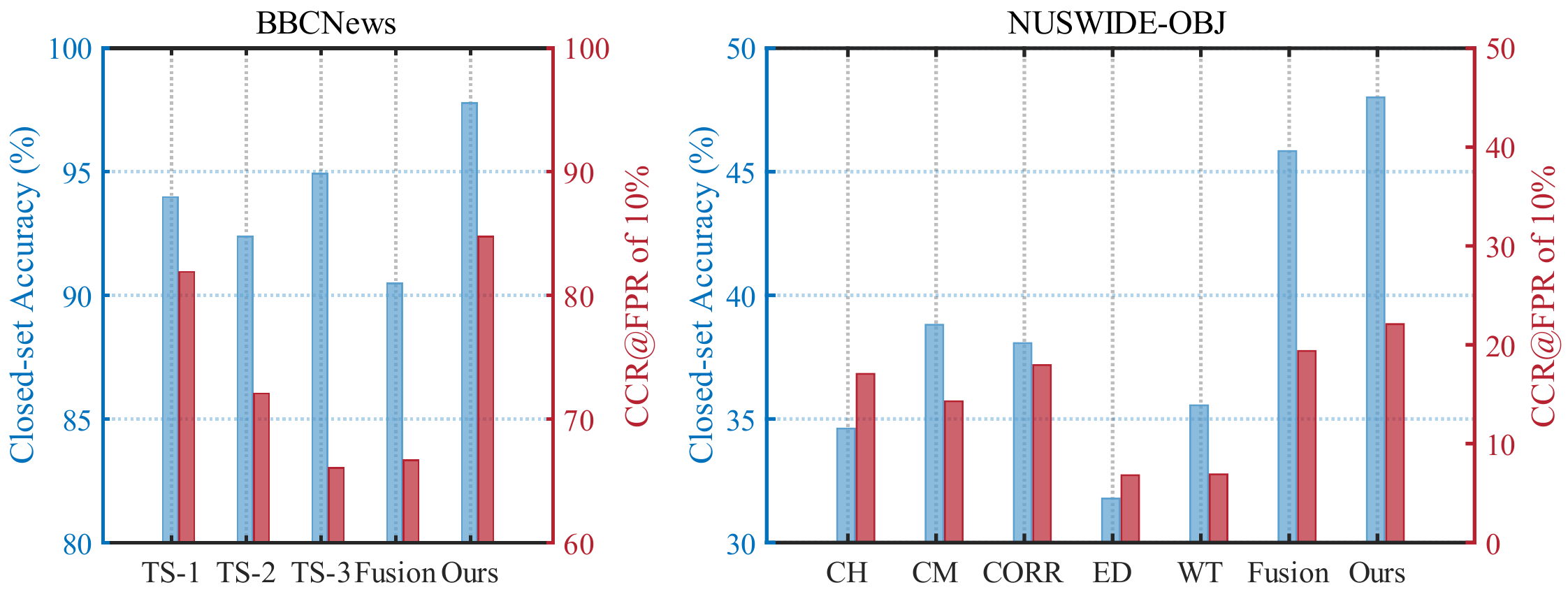}\\
  \caption{The impact of static view-induced bias on the performance of closed-set and open-set classification on different views of the BBCNews and NUSWIDE-OBJ datasets.
For clarity, we denote each text view of BBCNews as TS-1–TS-3 (Text Segments), and each visual feature of NUSWIDE-OBJ as CH (Color Histogram), CM (Color Moments), CORR (Color Correlation), ED (Edge Distribution), and WT (Wavelet Texture).
}
  \label{ACC}
\end{figure}
\begin{figure*}[h]
 \centering
 \includegraphics[width=\linewidth]{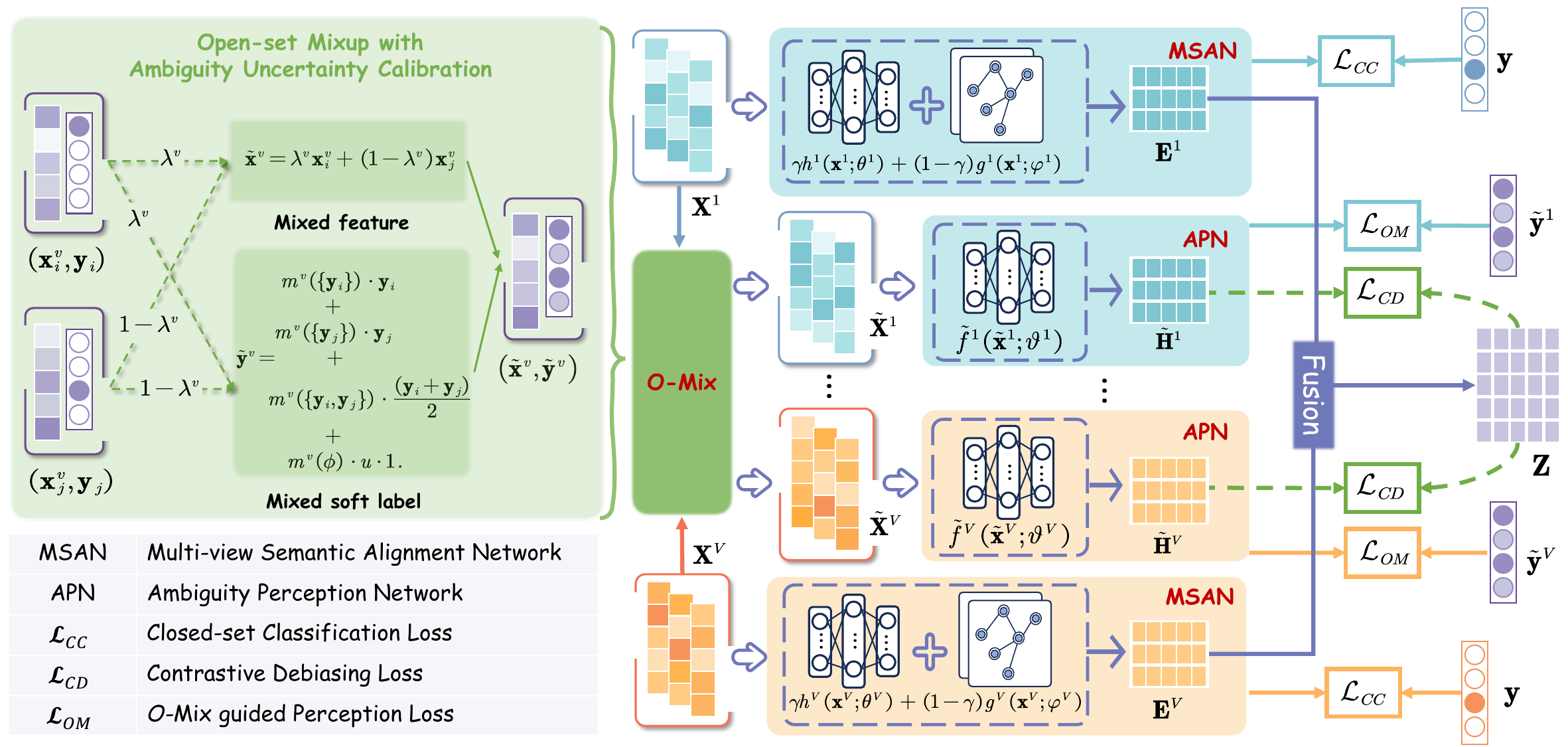}\\
 \caption{Three key components of the proposed framework: 
(1) MSAN extracts view-consistent representations from original inputs across all views; 
(2) \textit{O}-Mix simulates ambiguous samples by mixing view-specific instances and generating ambiguity-calibrated soft labels via DS-theory; 
(3) APN processes synthesized samples to model non-prototypical patterns. 
Overall, the three modules are jointly optimized to balance semantic alignment, ambiguity modeling, and view-wise debiasing.
}
 \label{framework}
\end{figure*}

Numerous problems in artificial intelligence involve making reliable predictions based on observations collected from diverse sources. 
A representative paradigm is multi-view data, also known as multi-modal data, which integrates heterogeneous attributes across different modalities, such as text, images, and other sensor signals~\cite{Chen22Adaptively,Jin23Deepalignment}. 
To effectively leverage this complementary information, researchers have developed multi-view learning methods to align heterogeneous feature spaces, thereby improving generalization across diverse tasks \cite{11027637, Wen23ScalableIncomplete}. 
However, most existing methods are trained on clean, complete datasets under the closed-world assumption that all test samples belong to known classes \cite{liangtrusted,wang2022adversarial, Chen24Concept}. 
Consequently, their reliability in realistic testing environments is limited, as they often misclassify ambiguous samples characterized by unclear class boundaries or conflicting modality prompts into known categories with unjustifiably high confidence \cite{10540001,luo2024fugnn}.

Therefore, multi-view learning needs to explore relevant solutions for open-set scenarios. 
In this context, two fundamental challenges emerge: 
 \textbf{unknown class recognition}  and  \textbf{static view-induced bias}.
 On one hand, the presence of unknown classes in unpredictable test environments demands that a multi-view learning model not only accurately classify known categories but also reliably identify unknown instances with appropriately low confidence \cite{du2024openviewer}.
On the other hand, static training data often exhibits view-induced bias, where different views of the same instance provide conflicting cues. 
This typically arises because the model has only observed a narrow slice of the view space, making it prone to overfit to view-specific patterns and further compounding the difficulty of consistent decision-making under uncertainty.
As a result, the model tends to establish spurious associations between certain views and class labels, undermining its ability to recognize unknown or ambiguous inputs~\cite{ke2023disentangling, zeng2024mitigating}.

%

Such biases are particularly problematic under open-set conditions, where models tend to rely on biased views rather than learning view-consistent representations.
This phenomenon is further corroborated by the empirical results shown in Figure~\ref{ACC}, where certain views (\textit{e.g.}, TS-3 and CM) achieve high accuracy under closed-set evaluation by exploiting static, view-specific features. 
 However, such features often fail to generalize to unknown categories, resulting in a noticeable drop in discriminative power under open-set conditions \cite{bao2024causal}.
This view-induced bias gives rise to misleading view–label associations, ultimately leading the model to produce overconfident yet incorrect predictions when encountering unfamiliar inputs.
In contrast, although TS-2 and CORR do not yield the highest closed-set performance, they produce more robust representations that better support the recognition of unknown classes.
 Moreover, directly fusing multi-view data may degrade the performance on unknown-class recognition, primarily because it tends to amplify the biases and noise present in individual views.
These discrepancies underscore the detrimental impact of static view-induced bias, wherein models fail to adequately capture representatives of the broader multi-view heterogeneous space, thereby hindering generalization in open-set scenarios.
These findings further motivate the design of a multi-view framework that exhibits higher predictive uncertainty when encountering unknown categories, characterized by more uniform and well-calibrated output distributions.
To address this, the proposed approach encourages the model to leverage a more diverse set of views, explicitly mitigates static view-induced bias, and promotes the integration of complementary, less biased representations for improved open-set generalization without losing closed-set performance.

In this paper, we propose a \textbf{M}ulti-view \textbf{O}pen-set learning framework via ambiguity uncertainty \textbf{C}alibration and view-wise \textbf{D}ebiasing (MOCD), which aims to gaining a broad perspective from multiple views, while ensuring unbiased recognition of new categories.  
The proposed method generates view-specific ambiguous representations through \textit{O}-Mix strategy, which models open-set uncertainty via Dempster–Shafer theory, incorporating calibration to enhance their reliability. 
To mitigate static view-induced biases, we introduce a contrastive debiasing module based on the Hilbert–Schmidt Independence Criterion (HSIC), which enforces statistical independence between ambiguous and view-consistent features.
This architecture ensures that the model benefits from diverse perspectives while preserving semantic consistency and resisting view-specific biases, ultimately enabling robust recognition of both known and unknown categories.
The architecture of MOCD is shown in Figure \ref{framework} and the contributions of this paper are as follows:
\begin{itemize}
\item  We introduce \textit{O}-Mix, a novel open-set sample synthesis strategy that generates virtual samples with calibrated ambiguity uncertainty, along with an auxiliary perception branch for open-set adaptation.

\item  We design a view-wise contrastive debiasing module that imposes independence constraints between view-sepecify ambiguous representations and consistent representations to suppress view-induced biases.

\item  Extensive experiments on diverse multi-view benchmarks demonstrate that MOCD consistently improves both known-class accuracy and unknown-class detection under open-set conditions. 

\end{itemize}

\section{Preliminaries}\label{sec:the}
\subsection{Generalized Uncertainty with Dempster–Shafer Theory}

Dempster–Shafer theory (DS-Theory) is a mathematical framework for modeling evidence and uncertainty.  
Let $\Theta = \{\theta_1, \theta_2, \dots, \theta_K\}$ denote the frame of discernment, \textit{i.e.}, the set of all known hypotheses or class labels.  
In DS-Theory \cite{sentz2002combination}, the basic event space can be extended to the power set $2^\Theta = \{\varnothing, \{\theta_1\}, \{\theta_1, \theta_2\}, \dots, \Theta\}$, where each subset corresponds to a possible (known or ambiguous) hypothesis.
A mass function, also known as a basic probability assignment (BPA), is a mapping $m: 2^\Theta \rightarrow [0,1] $, which satisfies
$$\sum_{A\in 2^\Theta}m(A)=1, m(\varnothing) = 0.$$
Here, $\varnothing$ is the null set and the value $m(A)$ represents the degree of belief assigned to the subset $A$ of $\Theta$.
However, in open-world settings, the frame of discernment is inherently incomplete, where unseen categories may appear at test time and cannot be explicitly enumerated within $\Theta$. 

Supposing that $U$ is a frame of discernment in an open setting, we adopt the Generalized Basic Probability Assignment (GBPA) framework \cite{Deng15}, which extends the standard BPA by allowing mass to be assigned to the unknown space.
\begin{Definition}
Generalized Basic Probability Assignment (GBPA) on a frame of discernment $U$,
with a mass function denoted as $m_G: 2_G^U \rightarrow [0, 1]$, which satisfies 
$$
\sum_{A\in 2^U_G}m_G(A)=1, m_G(\varnothing) \geq 0.
$$
\end{Definition}
The $\varnothing$ in GBPA is the focal element outside of the frame of discernment, distinct from the empty set in traditional BPA.
For simplicity, in the remainder of this paper, we have abbreviated the mass function $m_G(\cdot)$ to $m(\cdot)$.

\subsection{Data Augmentation}
In the context of data augmentation, Mixup extends the empirical training distribution by linearly interpolating between pairs of examples, which has been shown to improve the generalization ability of neural networks \cite{han2022umix,zhu2023openmix}.
Formally, given two random training examples $(\mathbf{x}_i, \mathbf{y}_i)$ and $(\mathbf{x}_j, \mathbf{y}_j)$, vanilla Mixup constructs a virtual training sample $(\tilde{\mathbf{x}}, \tilde{\mathbf{y}})$ by applying linear interpolation: 
\begin{equation}\label{genMix} 
\tilde{\mathbf{x}} = \lambda \mathbf{x}_i + (1 - \lambda) \mathbf{x}_j, \quad \tilde{\mathbf{y}} = \lambda \mathbf{y}_i + (1 - \lambda) \mathbf{y}_j, 
\end{equation} 
where the mixing coefficient $\lambda \in [0,1]$ is sampled from a Beta distribution with shape parameter $\tau$: $\lambda \sim \text{Beta}(\tau, \tau)$.
The interpolated sample $(\tilde{\mathbf{x}}, \tilde{\mathbf{y}})$ is then used to optimize the following objective: 
\begin{equation} \mathbb{E}_{{(\mathbf{x}_i, \mathbf{y}_i), (\mathbf{x}_j, \mathbf{y}_j)}} \left[ \ell\left( \tilde{\mathbf{x}}, \tilde{\mathbf{y}} \right) \right], 
\end{equation} 
where $\ell(\cdot)$ denotes the standard classification loss.
While traditional Mixup models interpolated uncertainty based on the proportion $\lambda$, it does not explicitly represent label ambiguity or uncertainty beyond known categories. 
To address this, DS-Mixup \cite{zhang2023new} combines Mixup with Dempster–Shafer theory, enabling more flexible modeling of belief over ambiguous label subsets. 
However, DS-Mixup still assumes a closed-world setting and fails to account for uncertainty induced by unknown categories in real-world open-set scenarios.
To this end, we adopt GBPA for explicitly modeling the uncertainty introduced by ambiguous mixtures in an open-set environment.

\section{The Proposed Framework}\label{sec:the}

\textbf{Problem Definition.}
In this paper, we consider the task of multi-view open-set classification, where the model is trained on a closed-set of 
$K$ known classes but may encounter samples from unknown classes during inference.
Given \(\mathcal{D} = \{\mathcal{X}_n, \mathbf{y}_n\}_{n=1}^N\),  each sample is composed of multiple views collected through different modalities or perspectives, formally represented as
\(\mathcal{X}_n = \{\mathbf{x}^{v}_n \in \mathbb{R}^{d_v} \}_{v=1}^V\).
The corresponding label $\mathbf{y}_n \in \{0, 1\}^K$  is a one-hot vector indicating the ground-truth class among the $K$ known categories.
Our target is to integrate information from multiple opinions to learn a sufficiently discriminative representation $\mathbf{z}_n \in \mathbb{R}^{D}$, from which the model produces a class probability vector $\mathbf{p}_n \in \mathbb{R}^{K}$.
This probability vector should not only assign probabilities to all known categories but also exhibit lower confidence when encountering unknown categories.

\subsection{Multi-view Semantic Alignment Network}

Each multi-view sample is associated with multiple distinct views, where each view contributes to accurate prediction.
To preserve view-specific characteristics while ensuring consistent semantics across views, features from different perspectives should be projected into semantically aligned representation spaces \cite{Liu24Samplelevel, Lyu24Common}.
Accordingly, we propose a Multi-view Semantic Alignment Network (MSAN) that allocates distinct processing modules to each view while aligning their outputs in a shared latent space.

\subsubsection{View-Specific Semantic Encoding.}

We first apply a multi-layer perceptron (MLP), denoted as $h(\cdot)$ , to extract high-level nonlinear transformations from raw view-specific features.
Formally, this process is expressed as: 
 \begin{equation} h(\mathbf{x};\theta): \mathbf{h} \leftarrow \sigma\left(\mathbf{W}^{(L)}\mathbf{h}^{(L-1)}+\mathbf{b}^{(L)} \right), \quad \mathbf{h}^{(0)}= \mathbf{x}, 
\end{equation}
 where $\theta=\{\mathbf{W}^{(l)}, \mathbf{b}^{(l)}\}_{l=1}^L$ 
denotes the set of trainable parameters, and $\sigma(\cdot)$ represents a nonlinear activation function (\textit{e.g.}, ReLU).
This hierarchical architecture enables the extraction of view-specific semantic patterns. For textual views, it may capture semantic topics or syntactic structures; while for visual views, it can extract edge or textures representations.

\subsubsection{Structure-Aware Semantic Refinement.}
Although $h(\cdot)$ independently processes the original features within a given view, it does not explicitly model relationships between samples.
 To incorporate such dependencies, we introduce $g(\cdot)$, which performs neighborhood aggregation to capture local structural relationships \cite{kipf2017semi}.
This operation is formally defined as: \begin{equation} 
g(\mathbf{x}_i;\varphi): \mathbf{m} \leftarrow \text{Update}(\mathbf{x}_i,\text{Agg}(\mathbf{x}_j| j \in \mathcal{N}(i))), 
\end{equation}
 where $\mathcal{N}(i)$ denotes the neighborhood of sample $\mathbf{x}_i$, and Agg$(\cdot)$ represents an aggregation function.
We employ the Clustering-with-Adaptive-Neighbors \cite{nie2014clustering} to dynamically select neighborhoods to capture both local and global data characteristics.

\subsubsection{Semantic Alignment Across Views}
For each view $v$,  the training of MSAN $f^{v}(\cdot)$ comprises two synergistic components tailored to preserve and transform view-relevant information.
The final view-specific representation $\mathbf{e}^{v} \in \mathbb{R}^{K}$  integrates both components
\begin{equation}\label{comRL}
f^{v}(\mathbf{x}^{v};\{\theta^{v},\varphi^{v}\}): \mathbf{e}^{v} \leftarrow \gamma  h^{v}(\mathbf{x}^{v};\theta^{v})+(1-\gamma) g^{v}(\mathbf{x}^{v};\varphi^{v}),
\end{equation}
where $\gamma$ is a learnable parameter that balances the contribution of feature extraction and structural preservation.
The learned representation $\mathbf{e}^{v}$ can be interpreted as an assignment of view-specific features into a unified latent space, not only considering individual information but also aggregating neighborhood features.
During training, we optimize parameters using a cross-entropy loss to ensure discriminative feature learning and maintain consistency across views:
\begin{equation}
\mathcal{L}_{\mathrm{CC}}^{v} = 
\mathbb{E}_{(\mathbf{x}^{v}_i,  \mathbf{y}_i)} \left[ \ell( f^{v}_{\{\theta^{v},\varphi^{v}\}}(\mathbf{x}_i^{v}), \mathbf{y}_i) \right].
\end{equation}
By jointly minimizing this closed-set classification loss across all views, the framework aligns heterogeneous feature spaces while preserving view-specific individuality, thereby enhancing the quality and coherence of the learned representations.

Given view-specific representations  $\{\mathbf{e}^{v}\}_{v=1}^V$, we compute the final view-consistent representation by averaging: 
\begin{equation} \label{fusion} \mathbf{z}=\frac{1}{V}\sum_{v=1}^V \mathbf{e}^{v}. 
\end{equation} 
Such uniform averaging avoids overfitting to dominant views and encourages balanced integration across all available modalities, which is especially beneficial in open-set scenarios where reliance on a single biased view can lead to unreliable predictions.

\subsection{\textit{O}-Mix: Open-set Mixup with Ambiguity Uncertainty Calibration}
To enhance the model’s adaptability to unknown samples, we introduce an auxiliary branch network trained on virtual samples, which are generated to simulate ambiguous or unknown cases.

\subsubsection{Modeling Ambiguity via Dempster–Shafer Theory}
To improve open-set recognition, we propose \textit{O}-Mix, a Mixup-based strategy that introduces intermediate ambiguous states and explicitly models the presence of unknown categories.
Firstly, we generate virtual samples in specific-view with vanilla Mixup:
\begin{equation}
\begin{array}{ll}\label{genOMix}
\tilde{\mathbf{x}}^{v}= \lambda^{v}\mathbf{x}_i^{v}+\left(1-\lambda^{v}\right){\mathbf{x}}_j^{v}.
\end{array}
\end{equation}
Then we define the hypothesis space for the mixed sample as $\Omega = \{\mathbf{y}_i, \mathbf{y}_j \}$,  where $\mathbf{y}_i$ and $\mathbf{y}_j$ are the ground-truth labels of two original samples.
This leads to the following mass constraints on the focal elements:
\begin{equation}
\left\{ \begin{array}{l}
m^{v}(\{\mathbf{y}_i, \mathbf{y}_j\})+ m^{v}(\{\mathbf{y}_i\})+m^{v}(\{\mathbf{y}_j\})+m^{v}(\varnothing)=1,\\
m^{v}(\{\mathbf{y}_i\})+m^{v}(\{\mathbf{y}_j\})=1-u.
\end{array} 
\right. 
\end{equation}
Here, $\{\mathbf{y}_i,\mathbf{y}_j\}$ is an ambiguous subset, and $m^{v}(\{\mathbf{y}_i, \mathbf{y}_j\})$  expresses the ambiguity when category information cannot be clearly assigned after sample mixing in this $v$-th view.
$m^{v}(\varnothing)$ represents general uncertainty outside the current identification scope, including potential unknown categories, and $u \in [0,1]$ determines the total mass assigned to ambiguous and unknown subsets.

Given a tunable ambiguity budget $a \leq u$, the mass values of the focal elements are calculated as:
\begin{equation}
\left\{ \begin{array}{l}
m^{v}(\{\mathbf{y}_i\})=\lambda^{v} \times (1-u),\\
m^{v}(\{\mathbf{y}_j\})=(1-\lambda^{v})\times (1-u),\\
m^{v}(\{\mathbf{y}_i,\mathbf{y}_j\})=a,\\
m^{v}(\varnothing)=u-a.\\
\end{array} 
\right. 
\end{equation}
In this way, \textit{O}-Mix generates ambiguous intermediate samples and calibrates their uncertainty through DS-Theory, enabling the model to explicitly calibrate both class ambiguity and open-set uncertainty.
Consequently, the soft label $\tilde{\mathbf{y}}^{v}$ corresponding to mixed sample $\tilde{\mathbf{x}}^{v}$ is defined as:
\begin{equation}
\begin{array}{ll}\label{genOMixlabel}
\tilde{\mathbf{y}}^{v}=m^{v}(\{\mathbf{y}_i\})\cdot \mathbf{y}_i+m^{v}(\{\mathbf{y}_j\}) \cdot \mathbf{y}_j\\
\ \ \ \ \ \ \ \ \ \ \ \ \ \  \ \ \ \ \ \ \  \ \ \ \  \ \ \ \ \ \ \ + m^{v}(\{\mathbf{y}_i, \mathbf{y}_j\})\cdot \frac{\mathbf{y}_i+\mathbf{y}_j}{2}+m^{v}(\varnothing)\cdot u \cdot \mathbf{1}.
\end{array}
\end{equation}
Here, $m^{v}(\varnothing)\cdot u \cdot \mathbf{1}$ denotes a uniform distribution over all known classes, ensuring that uncertainty is equally distributed.

\subsubsection{ Ambiguity Allocation for \textit{O}-Mixed Label}

To determine the optimal ambiguity split between $m^{v}(\{\mathbf{y}_i, \mathbf{y}_j\})$  and $m^{v}(\varnothing)$, we follow the principle of maximum uncertainty.
The entropy of the mass function with respect to $a$ is:
\begin{equation}
\begin{array}{l}
H(a)=-a \log(a)-(u-a) \log(u-a),
\end{array} 
\end{equation}
where we omit constant terms independent of $a$.
Take the derivative:
$
\frac{dH}{da}=-\log(a)-1+\log(u-a)+1.
$
Setting it to 0, we obtain $\log(a)=\log(u-a)$.
Thus, we allocate the uncertainty equally between the ambiguous subset and the unknown mass:
\begin{equation}
m^{v}(\{y_i,y_j\})=m^{v}(\varnothing)=\frac{u}{2},
\end{equation}
ensuring the most uniform uncertainty distribution among focal elements and avoiding bias toward either known ambiguity or unknown categories.

Furthermore, to avoid manual tuning of $u$, we introduce an adaptive scaling strategy based on the Mixup coefficient $\lambda^v$ in view $v$. 
Intuitively, $\lambda^v \approx 0.5$ indicates high ambiguity, warranting stronger uncertainty modeling. We define
$
u = c \cdot (1 - |\lambda^v - 0.5|),
$
where $c \in [0,1]$ is a scaling factor controlling the maximum uncertainty level.
This adaptive mechanism aligns the uncertainty modeling with the semantic ambiguity introduced by \textit{O}-Mix.

\subsection{View-wise Contrastive Debiasing}
Based on \textit{O}-Mix, we further design a debiasing loss to alleviate view-specific bias, facilitating learning more generalizable representations under open-set conditions.

\begin{algorithm}[t]
\caption{Training phase of the proposed framework}
\label{Algorithm}
\begin{algorithmic}[1]
\REQUIRE Multi-view training data $\mathcal{D}_{train}$, penalty coefficients $\alpha, \beta$, initial value of balance factor $\gamma$.
\STATE Initialize model parameters $\{\theta^{v}, \varphi^{v}, \vartheta^{v}\}_{v=1}^{V}$;
\FOR{\textbf{each} training epoch}
    \FOR{$v = 1$ to $V$}
        \STATE Extract view-specific representation $\mathbf{e}^{v}$ using MSAN as Eq.~\eqref{comRL};
        \STATE Generate a mixed sample $\{\tilde{\mathbf{x}}^v, \tilde{\mathbf{y}}\}$ using \textit{O}-Mix based on Eqs.~\eqref{genOMix} and \eqref{genOMixlabel};
        \STATE Compute ambiguity-aware representation $\tilde{\mathbf{h}}^{v}$ using APN  as Eq.~\eqref{ambresp};
    \ENDFOR
    \STATE Fuse $\{\mathbf{e}^{v}\}_{v=1}^V$ to obtain the view-consistent representation $\mathbf{z}$  as Eq.~\eqref{fusion};
    \STATE Update model parameters $\Theta$ by minimizing the total loss Eq.~\eqref{allloss};
\ENDFOR
\RETURN Optimized model parameters.
\end{algorithmic}
\end{algorithm}
\subsubsection{Auxiliary Branch for Uncertainty Perception}
These mixed samples are further regarded as substitutes for unknown classes, helping the model adapt to open-set scenarios.
To explicitly model the representation distribution of ambiguous samples, we introduce the lightweight Ambiguity Perception Network (APN), an auxiliary module dedicated to recognizing ambiguous cues:
\begin{equation}\label{ambresp}
\tilde{\mathbf{h}}^{v}=\tilde{f}^{v}(\tilde{\mathbf{x}}^{v};\vartheta^{v}),
\end{equation}
where $\tilde{f}^{v}(\cdot)$ denotes the branch encoder. 
This branch focuses on capturing atypical and uncertain patterns.
The \textit{O}-Mix guided perception loss for APN is defined as:
\begin{equation}
\begin{aligned}
\mathcal{L}_{\mathrm{OM}}^{v}=\mathbb{E}_{\{(\mathbf{x}_i^{v}, \mathbf{y}_i), (\mathbf{x}_j^{v}, \mathbf{y}_j)\}} \Big[  
\left(\lambda^{v} (1 - u) + \frac{u}{4} \right) \cdot \ell\left(\tilde{f}^{v}_{\vartheta^{v}}\left(\tilde{\mathbf{x}}^{v}\right), \mathbf{y}_i\right)  \\
 + 
\left((1 - \lambda^{v}) (1 - u) + \frac{u}{4} \right) \cdot \ell(\tilde{f}^{v}_{\vartheta^{v}}\left(\tilde{\mathbf{x}}^{v}), \mathbf{y}_j\right)+ \frac{u}{2} \cdot \ell\left(\tilde{f}^{v}_{\vartheta^{v}}\left(\tilde{\mathbf{x}}^{v}\right), \mathbf{1}\right) 
\Big].
\end{aligned}
\end{equation}
By combining Mixup-based sample generation with calibrated ambiguity uncertainty, \textit{O}-Mix enables the network to better capture class ambiguity and maintain cautious predictions on unfamiliar inputs. 
Furthermore, the auxiliary APN complements the main network by promoting uncertainty-calibrated feature learning, which is critical for reliable open-set recognition.

\subsubsection{Contrastive Debiasing Loss}
In open-set learning, static bias in the training data is a critical challenge that undermines generalization to unseen classes \cite{bao2021evidential,su2023hsic}. 
While in multi-view learning, different perspectives may contain biased view-label associations, where information is highly correlated with the category label but not universal (\textit{e.g.}, features such as color and texture in a certain perspective happen to be related to a certain category).

To ensure that the fused representation \( \mathbf{z} \) extracted by the backbone encoder captures consistent and generalizable semantics across views, we introduce a contrastive debiasing mechanism based on the Hilbert-Schmidt Independence Criterion (HSIC)~\cite{GrettonBSS05}.
 Specifically, we aim to suppress the dependency between the fused representation \( \mathbf{z} \) from MSAN and the view-specific ambiguous features \( \tilde{\mathbf{h}}^v \) extracted by APN.
 \cite{GrettonBSS05}. 
Formally, the contrastive debiasing loss is defined as:
\begin{equation}
\mathcal{L}^v_{CD}= \text{HSIC}(\mathbf{Z}, \tilde{\mathbf{H}}^v) = \frac{1}{(n - 1)^2} \text{Tr}(\mathbf{K}_Z \mathbf{C} \mathbf{K}_H \mathbf{C}),
\end{equation}
where $\mathbf{K}_Z$ and $\mathbf{K}_H$ are Gaussian kernels and $\mathbf{C} = \mathbf{I} - \frac{1}{n} \mathbf{1}\mathbf{1}^\top$ is the centering matrix. 
 On one hand, minimizing this dependency encourages the model to isolate view-specific biases in $\tilde{\mathbf{h}}^v$, while retaining a more general, shared semantic representation in $\mathbf{z}$.
On the other hand, reducing the dependency of $\mathbf{z}$ on static bias information can prevent the model from memorizing non-generalized features in the training set, which helps improve the recognition ability of unknown categories.

\subsection{Overall Objective}

The training objective of our framework consists of three complementary components:

\textbf{Closed-set Classification Loss} $\mathcal{L}_{CC}$: This loss supervises MSAN using labeled samples from known categories, ensuring discriminative performance under the closed-set setting.

\textbf{\textit{O}-Mix guided Perception Loss} $\mathcal{L}_{OM}$: Derived from \textit{O}-Mix strategy, this loss supervises APN using virtual samples with explicitly modeled open-set ambiguity uncertainty.

\textbf{Contrastive Debiasing Loss} $\mathcal{L}_{CD}$: Implemented via HSIC-based constraint, this loss isolates view-specific biases and enhances generalization to unknown categories.

The overall training loss is defined as:
\begin{equation}\label{allloss}
\mathcal{L}_{\text{total}} =\sum_{v=1}^V \mathcal{L}^v_{CC} + \alpha \mathcal{L}^v_{OM} + \beta \mathcal{L}^v_{CD},
\end{equation}
where $\alpha$ and $\beta$ are penalty coefficients that balance the contributions of the auxiliary branch and the debiasing term.

\textbf{Training and Inference.}
During training, we generate virtual samples using \textit{O}-Mix strategy in each minibatch and feed them into APN. 
 This approach not only simulates ambiguous samples during training, but also encourages the model to assign calibrated, less confident predictions when faced with inputs that fall outside the known class space.
Simultaneously, MSAN is optimized using a closed-set classification loss on clean samples, and regularized by a contrastive debiasing loss to mitigate view-specific biases.
The training phase is summarized in Algorithm \ref{Algorithm}.

During inference, given test samples $\{\mathbf{x}^{v}_*\}_{v=1}^V$, the representation $\mathbf{z}_*$ is computed by aggregating MSAN outputs across all views: 
\begin{equation}
\mathbf{z}_* = \frac{1}{V}\sum_{v=1}^V \left( f^{v}(\mathbf{x}_*^{v};\{\theta^{v}, \varphi^{v}\}) \right),
\end{equation}

\begin{table*}[t]
\centering
\caption{CCR at different FPR for all compared algorithms under \textit{Openness}=0.1 setting. The best and the second best values are highlighted in \textcolor{red}{red} and \textcolor{blue}{blue}, respectively.}
\resizebox{0.92\textwidth}{!}{
\begin{tabular}{c|ccccc|ccccc|cccccccc}
\toprule
Methods$\backslash$Datasets  & \multicolumn{5}{c}{BBCNews} & \multicolumn{5}{c}{Caltech20} & \multicolumn{5}{c}{Hdigit}\\ 
\cmidrule(r){2-6} \cmidrule(r){7-11} \cmidrule(r){12-16}    
CCR@FPR of  & 1.0\%&5.0\%&10\%&50\%&100\% & 1.0\%&5.0\%&10\%&50\% &100\% & 1.0\%&5.0\%&10\%&50\% &100\%\\
\midrule

CoGCN & 40.11 & \textcolor{blue}{58.76} & \textcolor{blue}{62.99} & \textcolor{blue}{75.14} & \textcolor{blue}{80.51} & 60.34 & 67.65 & 71.54 & 79.99 & 84.66 & 61.36 & 84.21 & 89.91 & 98.2 & 99.29 \\
MMDynamics & 44.07 & 54.24 & 59.04 & 70.90 & 78.81 & 57.16 & 75.49 & \textcolor{blue}{80.05} & \textcolor{blue}{87.72} & \textcolor{blue}{90.47} & 77.09 & 88.75 & 92.57 & 98.45 & 99.36 \\
MvNNcor & 35.31 & 47.74 & 55.93 & 73.73 & 77.40 & \textcolor{blue}{67.23} & \textcolor{blue}{76.03} & 79.57 & 85.68 & 88.38 & 72.79 & 85.79 & 89.80 & 97.20 & 98.84 \\
TMC & 39.55 & 46.89 & 48.87 & 54.52 & 66.10 & 31.82 & 48.89 & 61.17 & 75.25 & 81.49 & 65.61 & 76.79 & 82.91 & 95.02 & 97.82 \\
RCML & 37.57 & 46.33 & 52.54 & 65.25 & 74.86 & 34.33 & 44.88 & 57.70 & 74.60 & 82.14 & 74.32 & 86.00 & 89.64 & 96.95 & 98.82 \\
MAMC & 7.34 & 24.58 & 31.64 & 46.33 & 50.85 & 11.50 & 13.42 & 15.46 & 41.64 & 56.14 & 55.50 & 78.38 & 84.91 & 96.93 & 98.77 \\
TUNED & \textcolor{blue}{44.63} & 55.93 & 58.19 & 64.69 & 76.24 & 56.86 & 67.23 & 72.20 & 81.37 & 85.68 & \textcolor{blue}{88.43} & \textcolor{blue}{95.21} & \textcolor{blue}{96.96} & \textcolor{blue}{99.20} & \textcolor{blue}{99.61} \\
Ours & \textcolor{red}{55.08} & \textcolor{red}{66.10} & \textcolor{red}{69.49} & \textcolor{red}{81.36} & \textcolor{red}{88.70} & \textcolor{red}{71.78} & \textcolor{red}{81.67} & \textcolor{red}{84.54} & \textcolor{red}{89.69} & \textcolor{red}{92.33} & \textcolor{red}{90.43} & \textcolor{red}{96.25} & \textcolor{red}{97.89} & \textcolor{red}{99.61} & \textcolor{red}{99.89} \\

\midrule \midrule

Methods$\backslash$Datasets  & \multicolumn{5}{c}{ Iaprtc12 } & \multicolumn{5}{c}{NUSWIDE-OBJ}& \multicolumn{5}{c}{VGGFace2}\\\cmidrule(r){2-6} \cmidrule(r){7-11} \cmidrule(r){12-16}  
CCR@FPR of  & 1.0\%&5.0\%&10\%&50\%&100\% & 1.0\%&5.0\%&10\%&50\% &100\% & 1.0\%&5.0\%&10\%&50\% &100\%\\
\midrule
CoGCN & \textcolor{red}{15.05} & \textcolor{blue}{26.89} & \textcolor{blue}{33.13} & 61.10 & 75.46 & 2.97 & 7.440 & 10.97 & 26.55 & 36.34 & 8.570 & 18.47 & 23.51 & 39.85 & 46.90 \\
MMDynamics & 2.330 & 6.050 & 12.46 & 50.24 & 74.30 & 3.22 & 7.940 & 12.11 & 29.18 & 39.02 & 10.62 & 18.78 & 23.23 & 37.67 & 44.14 \\
MvNNcor & 10.70 & 18.27 & 23.18 & 50.60 & 71.72 & 2.77 & 8.110 & 12.15 & 30.39 & 41.55 & 9.700 & 18.16 & 23.07 & 38.15 & 45.36 \\
TMC & 5.290 & 13.94 & 20.70 & 54.03 & 74.87 & 4.63 & 10.16 & 14.23 & 31.37 & 40.95 & 12.21 & 20.06 & 24.50 & 37.92 & 44.29 \\
RCML & 5.310 & 12.98 & 20.30 & 53.31 & 75.11 & 3.82 & 8.590 & 13.23 & 29.42 & 37.98 & 9.970 & 17.45 & 21.32 & 34.22 & 40.17 \\
MAMC & 1.410 & 5.530 & 13.32 & 53.31 & 75.68 & 5.50 & \textcolor{blue}{12.91} & 16.94 & 32.75 & \textcolor{blue}{42.08} & 12.06 & 20.34 & 25.15 & 38.56 & 45.00 \\
TUNED & 11.62 & \textcolor{red}{27.04} & 29.85 & \textcolor{blue}{63.82} & \textcolor{blue}{81.42} & \textcolor{blue}{6.10} & 12.76 & \textcolor{blue}{17.43} & \textcolor{blue}{32.78} & 41.33 & \textcolor{blue}{15.36} & \textcolor{blue}{24.80} & \textcolor{blue}{29.97} & \textcolor{blue}{45.48} & \textcolor{blue}{52.76} \\
Ours & \textcolor{blue}{14.32} & 25.89 & \textcolor{red}{33.54} & \textcolor{red}{67.77} & \textcolor{red}{85.87} & \textcolor{red}{9.10} & \textcolor{red}{16.67} & \textcolor{red}{21.58} & \textcolor{red}{37.47} & \textcolor{red}{47.07} & \textcolor{red}{27.40} & \textcolor{red}{39.53} & \textcolor{red}{45.57} & \textcolor{red}{62.10} & \textcolor{red}{69.40} \\
\bottomrule
\end{tabular}}\label{CCRClassification}
\end{table*}
\begin{table}[t]
\renewcommand\arraystretch{1}
\centering
\caption{Statistics of the test multi-view datasets.}
\label{multiDataDescription}
\resizebox{0.48\textwidth}{!}{
\begin{tabular}{lllll}
\toprule
Dataset & \#Samples & \#Feature Dimensions& \#Classes & Types \\
\midrule
BBCNews   &685 &4,659/4,633/4,665&5   & Text Documents \\
Caltech20  &2,386&48/40/254/1,984/512/928&20&Object Recognition \\
Hdigit     & 10,000&784/256&10&   Digit Images \\
Iaprtc12    &7,855 &100/100&6& Image-Text Pairs \\
NUSWIDE-OBJ&30,000&65/226/145/74/129&31& Object Recognition \\
VGGFace2    & 34,027&944/576/512/640/50&50&Face Images \\
\bottomrule
\end{tabular}}
\end{table}

\section{Experimental Results and Study}

\subsection{Experimental Settings}

\subsubsection{Datasets and Comparative Methods}
We cover six datasets with multiple views, the statistics are given in Table \ref{multiDataDescription}.
We also compare the classification performance of MOCD with the following deep multi-view models, including CoGCN~\cite{li2020co}, MvNNcor~\cite{xu2020deep}, and MAMC~\cite{Lin25Enhance}, as well as uncertainty-based methods such as MMDynamics~\cite{han2022multimodal}, TMC~\cite{HanZFZ21}, RCML~\cite{xu2024reliable}, and TUNED~\cite{huang2024trusted}.

\subsubsection{Evaluation Metric}
Referring to the evaluation metric in the Open-Set Classification Rate (OSCR) \cite{dhamija2018reducing}, we compute the Correct Classification Rate (CCR) versus False Positive Rate (FPR) to assess model performance in open-set scenarios.
Given a score threshold $p$, FPR and CCR are defined as
\begin{equation}
\begin{array}{ll}\label{genMix}
\text{FPR}(p) = \frac{\left| \{ x \mid x \in \mathcal{D}_u, \max\limits_{c} P(c \mid x) \geq p \} \right|}{|\mathcal{D}_u|},\\
\text{CCR}(p) = \frac{\left| \{ x \mid x \in \mathcal{D}_c, \arg\max\limits_{c} P(c \mid x) = \hat{c} \land P(\hat{c} \mid x) > p \} \right|}{|\mathcal{D}_c|},
\end{array}
\end{equation}
where \( \mathcal{D}_u \) and  \( \mathcal{D}_c \) represent the set of unknown samples and known-class samples, and \( \hat{c} \) is the correct class label. 
By varying \( p \), FPR indicates the proportion of unknown samples incorrectly classified as known classes, and CCR measures the fraction of correctly classified known samples.
An ideal model should maintain a high CCR at low FPR, ensuring accurate classification of known classes while effectively detecting unknown classes.

\subsubsection{Implementations}
Since the aforementioned algorithm is not designed for open environments, we adopt the same experimental settings, training in a closed-set environment while testing in an open-set environment. 
The dataset is split into a training set, validation set, and test set for known categories with the ratio of 10\%/10\%/80\%.  
We utilize openness to represent the complexity of the open-set task, as follows
$
\textit{Openness} = 1 - \sqrt{2 |\mathcal{D}_c|/(2|\mathcal{D}_c|+  |\mathcal{D}_u|)},
$
 A higher openness value indicates a greater proportion of unknown categories in the test set.
Our experiments are conducted with the default setting of \textit{Openness}=10\%.
We set the shape parameter of the Beta distribution to $\tau = 1$ for mixup-based sample generation.
The model is trained using a batch size of 64 and a learning rate of 0.003.
The balance factor $\gamma$ in Eq.~\eqref{comRL} is set to 0.7 to balance feature extraction and structural refinement.
Unless otherwise specified, the loss balancing coefficients are set to $\alpha = 1$ and $\beta = 1$.
All experiments are implemented by PyTorch on NVIDIA 3090 with 24GB memory.

\begin{figure*}[t]
  \centering
  \includegraphics[width=0.9\textwidth]{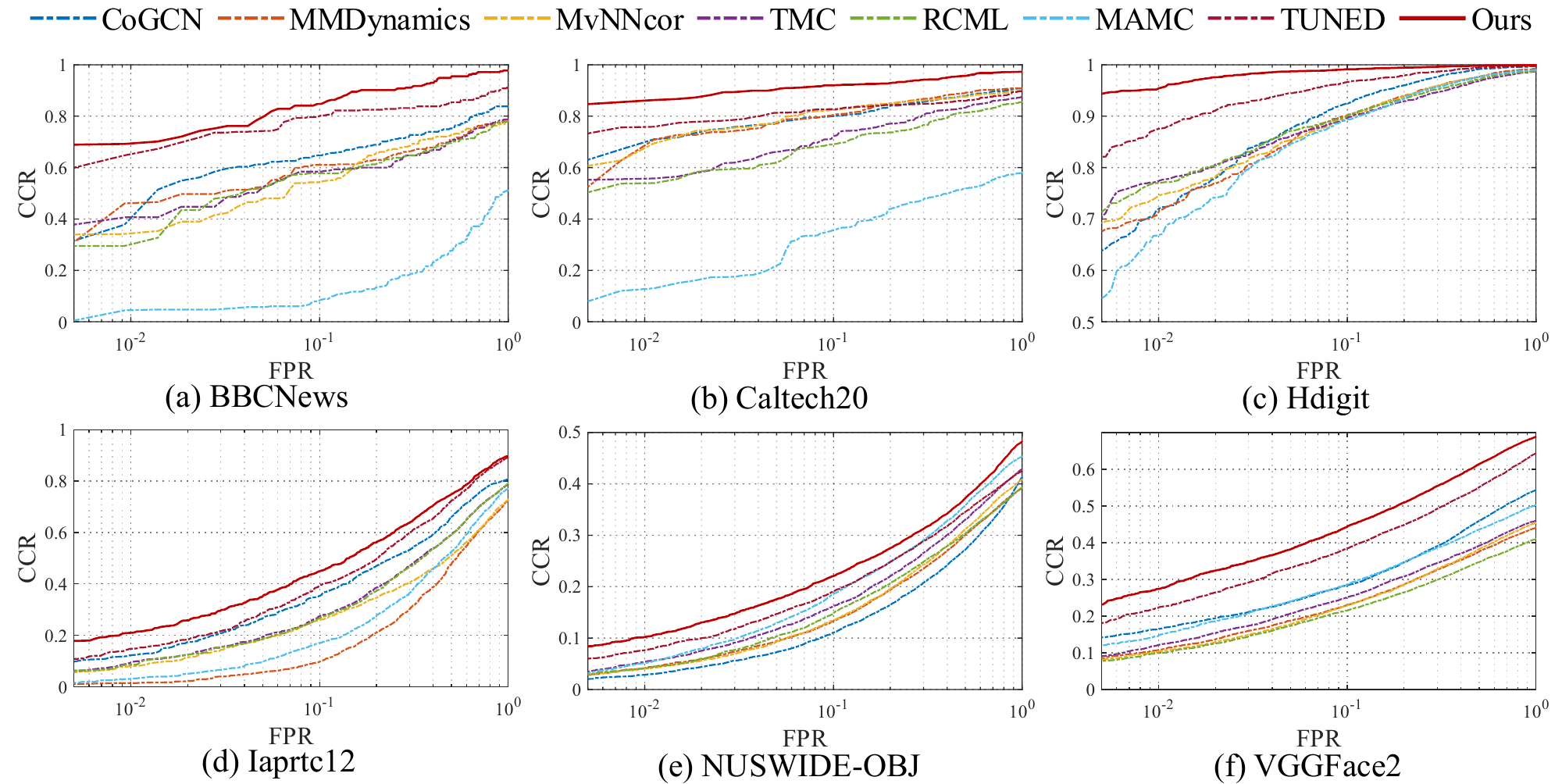}\\
  \caption{OSCR curves plotting the CCR over the FPR on all test multi-view datasets for all compared methods.}
  \label{ccrfpr}
\end{figure*}
\begin{figure*}[t]
  \centering
  \includegraphics[width=0.9\textwidth]{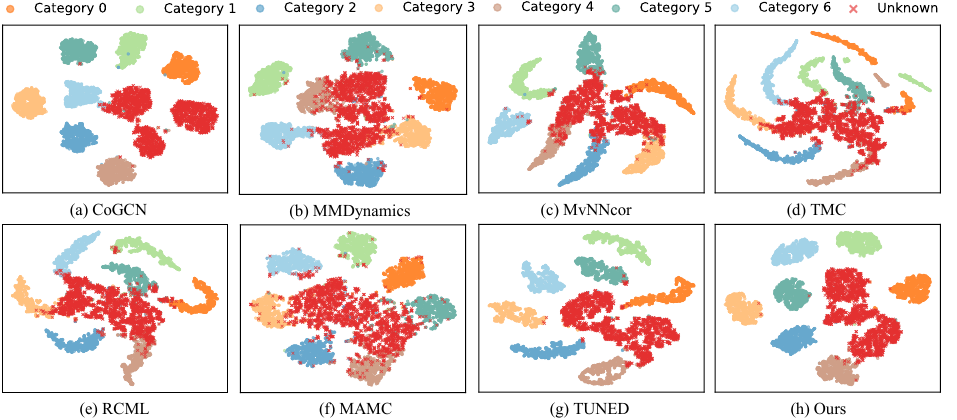}\\
  \caption{Visualization of representations with t-SNE learned by different methods on the Hdigit dataset.}
  \label{tsne}
\end{figure*}

\subsection{Overall Performance}
In this subsection, we compare our method against state-of-the-art multi-view classification techniques. 
It is important to note that when \( \text{FPR} = 100\% \), the CCR is solely determined by the fraction of correctly classified known samples, which is equivalent to the closed-set classification accuracy.
We provide a comprehensive evaluation and have the following observations:
(1) As shown in Table \ref{CCRClassification}, most models perform well in closed-set classification. 
Still, as the FPR decreases, indicating an increasing requirement for unknown class rejection, the performance of most models declines significantly. 
The proposed method retains a significant advantage in terms of lower FPR.
This highlights the challenge of balancing classification accuracy with open-set recognition capability.
(2) In Figure \ref{ccrfpr}, the performance gap widens at lower FPR values, suggesting stronger unknown class rejection capability.
For Iaprtc12 and NUSWIDE-OBJ, all models decline more sharply as FPR decreases, indicating a more challenging dataset for open-set recognition.
For Caltech20 and VGGFace2, the performance gap between models is more noticeable, with the red curve maintaining a clear advantage.
TMC performs the worst in most cases, indicating poor adaptability to open-set scenarios.
These results demonstrate the robustness of the proposed MOCD in multi-view learning and its superior capability for open-set recognition compared to existing baselines.
(3)
To intuitively investigate the distribution of known categories and unknown samples, we visualize the learned multi-view representation in Figure \ref{tsne}. 
All methods demonstrate the ability to form distinct clusters for known categories.
 However, the handling of unknown samples varies significantly across different approaches. 
MMDynamics and MvNNcor exhibit considerable overlap between known and unknown samples. 
For other methods, unknown samples are scattered throughout the feature space, indicating weaker rejection capabilities.
In contrast, the proposed method achieves the most structured feature representation, characterized by compact known-class clusters and clear separation from unknown samples.

%
%

\begin{figure}[t]
  \centering
  \includegraphics[width=0.48\textwidth]{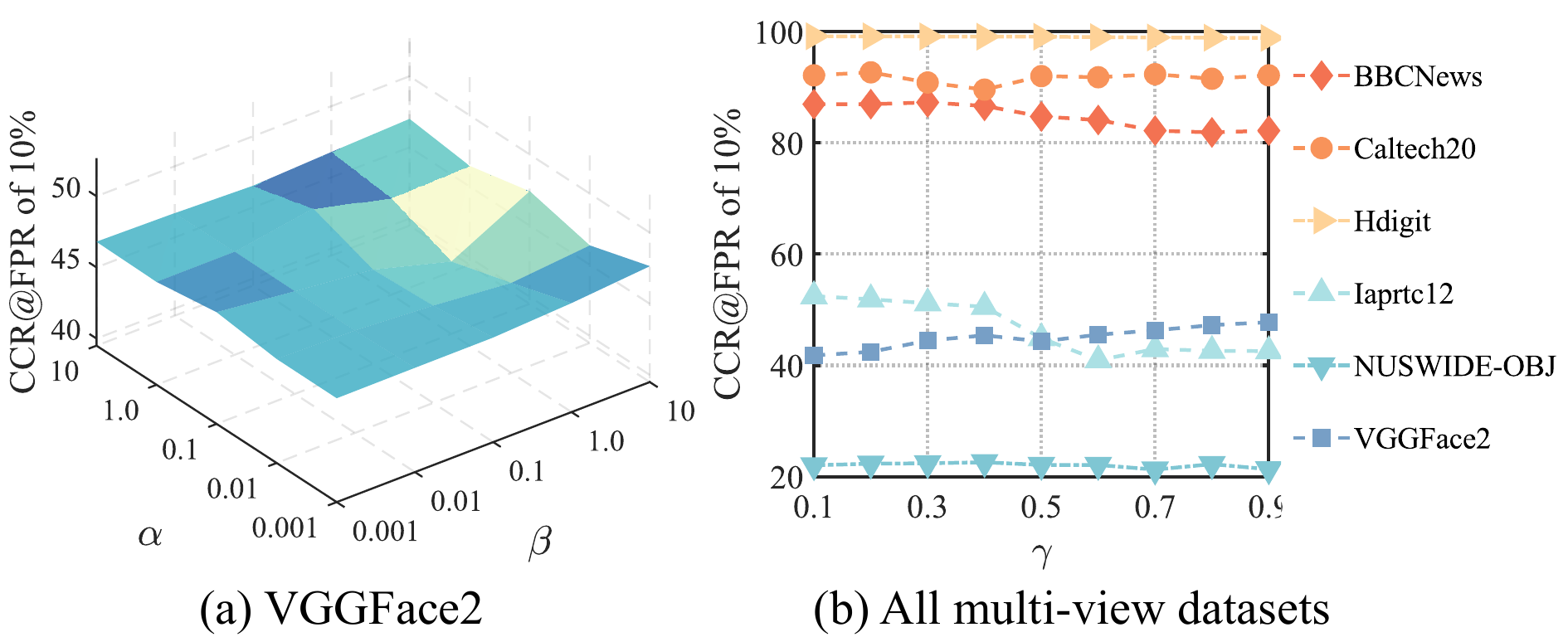}\\
  \caption{(a) Parameter sensitivity analysis of $\alpha$ and $\beta$ in proposed method on VGGFace2 datasets. (b) Parameter sensitivity analysis of $\gamma$ in the proposed method on all test datasets.}
  \label{alphabeta}
\end{figure}

\subsection{Parameter Analysis}
\subsubsection{Impact of Penalty Xoefficients $\alpha$ and $\beta$.}
We search $\alpha$ and $\beta$ in $[0.001, 0.01, \dots, 10]$.
As shown in Figure \ref{alphabeta}(a), on VGGFace2, CCR@FPR=10\% remains stable when both $\alpha$ and $\beta$ are within the lower to moderate range.
This suggests that while the auxiliary perception loss is robust to hyperparameter variations, excessive debiasing may suppress discriminative view-specific cues, particularly in more challenging datasets.

\subsubsection{Impact of Balance Factor $\gamma$.}
We vary the value of $\gamma$ from 0.1 to 0.9 to evaluate its impact on model performance. As shown in Figure~\ref{alphabeta}(b), our method consistently maintains stable performance across all datasets, demonstrating strong robustness to the choice of $\gamma$ and eliminating the need for precise tuning. Notably, most datasets achieve high CCR@FPR=10\% when $\gamma$ falls within the range of 0.5 to 0.7, indicating that a balanced contribution between feature extraction and structural preservation is beneficial for comprehensive representation learning.

\begin{figure}[t]
  \centering
  \includegraphics[width=0.42\textwidth]{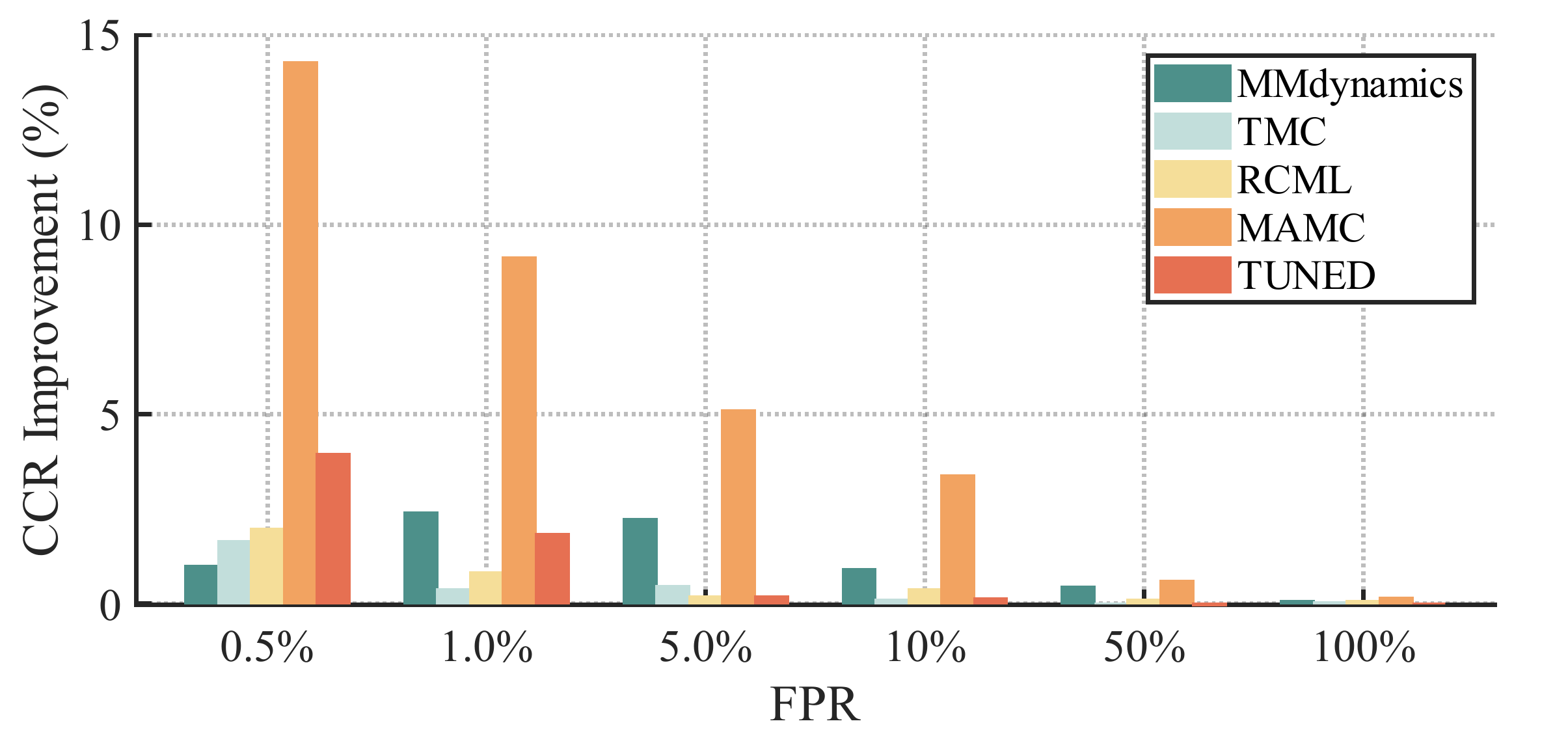}\\
  \caption{CCR improvement of \textit{O}-Mix as a plug-in compared to the original method on the Hdigit dataset at different FPRs.}
  \label{hdigitomixgain}
\end{figure}


\subsection{Further Analysis}
\subsubsection{Module Analysis.}
To further evaluate the contribution of \textit{O}-Mix, we conduct the ablation study in comparison with vanilla Mixup and \textit{O}-Mix.
From Table \ref{ablation}, we observe a clear performance improvement as each component is added. Starting from the baseline with only $h^{v}(\cdot)$, incorporating $g^{v}(\cdot)$ and standard Mixup provides moderate gains. 
The improvement over vanilla Mixup further validates the importance of explicitly modeling uncertainty and open-set ambiguity, rather than relying on naive interpolation alone.

\begin{table}[t]
\renewcommand\arraystretch{1}
\centering
\caption{Ablation study of each component evaluated by CCR@FPR of 10\% on all test multi-view datasets.}
\resizebox{0.45\textwidth}{!}{
\label{ablation}
\begin{tabular}{cccc|ccccccc}
\toprule
$h^{v}(\cdot)$ & $g^{v}(\cdot)$&Mixup&\textit{O}-Mix&BBCNews&Caltech20&Hdigit\\\midrule
\CheckmarkBold& \XSolidBrush& \XSolidBrush& \XSolidBrush&49.15& 56.44&89.73\\
 \CheckmarkBold& \CheckmarkBold& \XSolidBrush& \XSolidBrush&  71.47& 81.43&  96.57 \\
\CheckmarkBold& \CheckmarkBold& \CheckmarkBold & \XSolidBrush&70.34&86.64&97.61 \\
\CheckmarkBold& \CheckmarkBold & \XSolidBrush& \CheckmarkBold &69.49&84.54&97.89\\\midrule \midrule
$h^{v}(\cdot)$ & $g^{v}(\cdot)$&Mixup&\textit{O}-Mix&Iaprtc12&NUSWIDE-OBJ&VGGFace2\\\midrule
\CheckmarkBold& \XSolidBrush& \XSolidBrush& \XSolidBrush&23.13&17.94&22.52\\
 \CheckmarkBold& \CheckmarkBold& \XSolidBrush& \XSolidBrush&32.11&19.52&43.72\\
\CheckmarkBold& \CheckmarkBold& \CheckmarkBold & \XSolidBrush&32.32&20.14&  46.41 \\
\CheckmarkBold& \CheckmarkBold & \XSolidBrush& \CheckmarkBold &33.54&21.58& 45.57\\
\bottomrule
\end{tabular}}
\end{table}
\begin{figure}[t]
  \centering
  \includegraphics[width=0.46\textwidth]{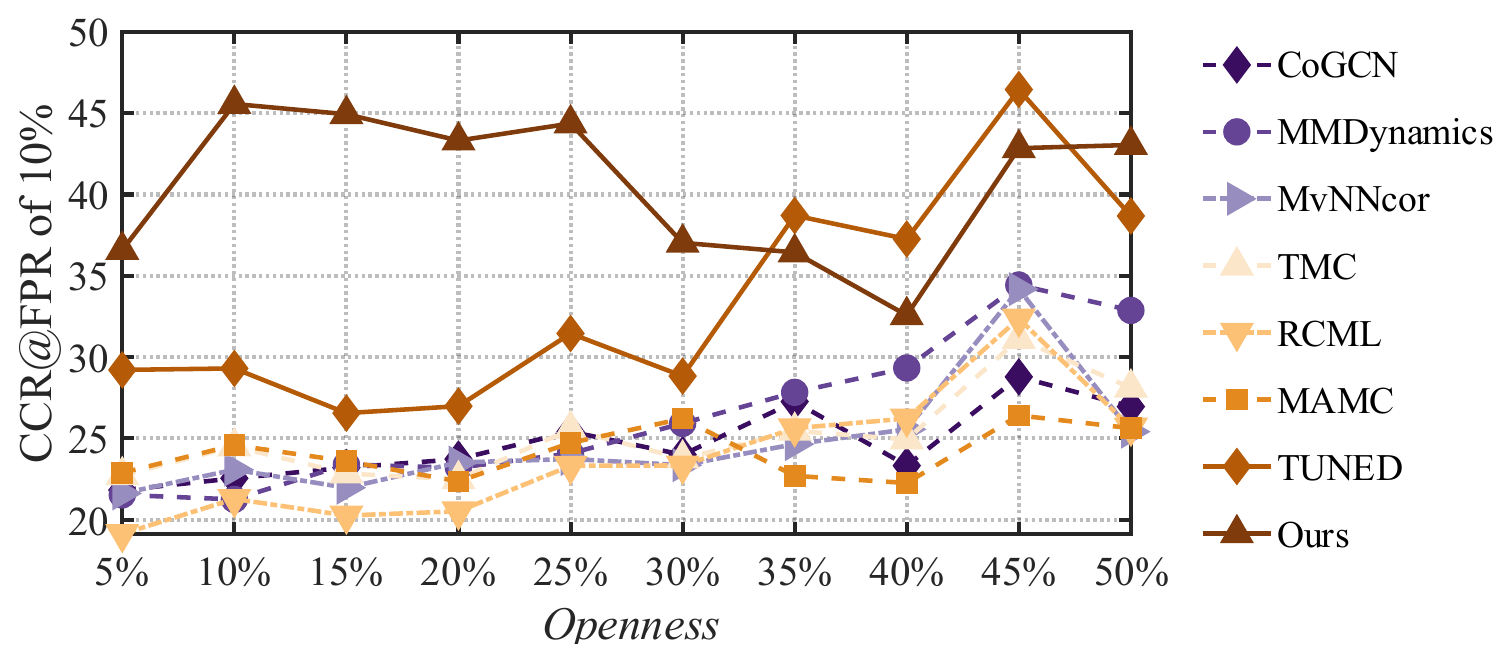}\\
  \caption{CCR@FPR of 10\% for all compared method on the VGGFace2 dataset under different \textit{Openness} setting.}
  \label{openness}
\end{figure}
\subsubsection{ Compatibility Study.}
We conduct a plug-and-play evaluation by integrating \textit{O}-Mix as an auxiliary module into several baseline methods. 
As shown in Figure~\ref{hdigitomixgain}, the improvements are most significant at lower FPRs (\textit{e.g.}, 0.5\% and 1\%), and gradually diminish as the threshold increases, becoming marginal at 5.0\% and 10\%. 
These results demonstrate that \textit{O}-Mix effectively enhances the model’s ability to recognize unknown classes without degrading performance on closed-set classification, highlighting its compatibility and generalization capability across diverse architectures.


\subsubsection{Openness Evaluation.}
We conduct an openness evaluation to assess how each method performs under increasing proportions of unknown classes.
Figure \ref{openness} demonstrates the robustness and effectiveness of our method under increasing open-set conditions. 
The consistently superior CCR@FPR=10\% across different openness levels indicates the generalization ability of our method. 

\section{Conclusion}
In this work, we proposed a multi-view open-set classification framework, namely MOCD, to improve generalization to unknown categories across multiple views.
To better capture the inherent ambiguity of virtual samples, we introduced \textit{O}-Mix, a Mixup-based strategy augmented with Dempster–Shafer theory, which incorporated mass assignments to model open-set uncertainty in an interpretable manner.
By linearly combining samples across classes and views, \textit{O}-Mix generated ambiguous data that disrupts view-induced biased features.
To suppress the influence of such biases, we enforced independence between the auxiliary branch output and the unified multi-view representation, encouraging the model to focus on more essential and generalizable semantics.
Overall, by promoting cross-view consistency while suppressing spurious view-label associations, the proposed method enables more inclusive recognition of unknown categories and consistently improves performance on both closed-set and open-set benchmarks.

\begin{acks}
This work is in part supported by the National Natural Science Foundation of China under Grants U21A20472 and 62276065, the Fujian Provincial Natural Science Foundation of China under Grant 2024J01510026, and the Fujian Provincial Department of Education Youth Project of China under Grant JZ230011.
\end{acks}

\bibliographystyle{ACM-Reference-Format}
\balance
\bibliography{ML}

\end{document}